\UseRawInputEncoding
\documentclass[letterpaper, 10 pt, journal, twoside]{IEEEtran} 

\title{ADMM-Based Safety-Critical Distributed NMPC for Cooperative Transportation by Quadrupedal Robots}

\usepackage{amsmath,amssymb,amsfonts}
\usepackage{cite}
\usepackage{graphicx} 
\usepackage{epsfig} 
\usepackage{times} 
\usepackage{xcolor}
\usepackage{balance}
\usepackage{multicol}
\usepackage{array}
\usepackage{multirow}
\usepackage{booktabs}
\usepackage[colorlinks=true, allcolors=blue]{hyperref}

\newcommand{\Real}{\mathbb{R}}
\newcommand{\col}{\textrm{col}}
\newcommand{\Integer}{\mathbb{Z}_{\geq0}}
\newcommand{\Integerpos}{\mathbb{Z}_{>0}}
\newcommand{\diag}{\textrm{diag}}
\newcommand{\bdiag}{\textrm{block diag}}
\newcommand{\identity}{\mathbb I}

\newcommand{\reff}{\mathrm{ref}}

\newcommand{\Graphnet}{\mathcal{G}}
\newcommand{\Vset}{\mathcal{V}}
\newcommand{\Aset}{\mathcal{A}}
\newcommand{\Eset}{\mathcal{E}}
\newcommand{\Xset}{\mathcal{X}}
\newcommand{\Uset}{\mathcal{U}}
\newcommand{\Oset}{\mathcal{O}}

\newcommand{\Zset}{\mathcal{Z}}
\newcommand{\Load}{\textrm{L}}
\newcommand{\SRB}{\textrm{SRB}}

\newcommand{\ex}{\textrm{e}}

\DeclareMathOperator*{\argmin}{arg\,min}

\usepackage{etoolbox}

\makeatletter
\patchcmd{\@makecaption}{\scshape}{}{}{}
\makeatother

\AtBeginEnvironment{equation*}{\vspace{-3pt}}
\AtEndEnvironment{equation*}{\vspace{-3pt}}
\AtBeginEnvironment{equation}{\vspace{-3pt}}
\AtEndEnvironment{equation}{\vspace{-3pt}}
\AtBeginEnvironment{align}{\vspace{-3pt}}
\AtEndEnvironment{align}{\vspace{-3pt}}
\AtBeginEnvironment{alignat}{\vspace{-3pt}}
\AtEndEnvironment{alignat}{\vspace{-3pt}}

\newtheorem{remark}{\textbf{Remark}}

\author{Ruturaj S. Sambhus$^{1}$, Kapi Ketan Mehta$^{1}$, Yicheng Zeng$^{1}$, and Kaveh Akbari Hamed$^{1}$

\thanks{The work of R.~S.~Sambhus and K.~Akbari Hamed is partially supported by the National Science Foundation (NSF) under Grant 2423725.}
\thanks{$^{1}$R.~S.~Sambhus, K.~K.~Mehta, Y.~Zeng, and K.~Akbari Hamed (\textit{Corresponding Author}) are with the Department of Mechanical Engineering, Virginia Tech, Blacksburg, VA 24061, USA, {\tt\small \{ruturajsambhus, kmehta4, zyicheng, kavehakbarihamed\}@vt.edu}}
}

\begin{document}
\maketitle


\begin{abstract}
This paper presents a safety-critical distributed nonlinear model predictive control (DNMPC) framework for cooperative payload transportation by teams of quadrupedal robots. The proposed approach models the robotic team and the shared payload as a dynamically coupled networked system with rigid holonomic coupling constraints arising from cooperative transportation. To enable distributed real-time optimization, the centralized finite-horizon optimal control problem is decomposed into parallel local NMPC subproblems coordinated through the alternating direction method of multipliers (ADMM). The resulting distributed framework enforces consensus over both payload-state and interaction-wrench trajectories while explicitly incorporating acceleration-level holonomic coupling constraints within the distributed predictive control formulation. Safety-critical obstacle avoidance constraints for both the robotic agents and payload are enforced using higher-order control barrier functions (HOCBFs). The framework is validated through numerical simulations with teams of two, three, and four quadrupedal robots transporting shared payloads in cluttered environments. Real-time experiments on two- and three-robot teams demonstrate safe and robust transportation under payload uncertainty and external disturbances. Compared with centralized NMPC, the proposed framework achieves up to $23\%$ reduction in average NLP solve time while maintaining comparable closed-loop performance. Ablation studies further demonstrate robustness to communication delays and show that explicit payload-state consensus and holonomic constraints substantially improve payload tracking and distributed coordination over existing wrench-only consensus formulations.
\end{abstract}

\begin{IEEEkeywords}
Legged robots, motion control, distributed robot systems
\end{IEEEkeywords}


\vspace{-1em}
\section{Introduction}
\label{sec:Intro}

Cooperative payload transportation using teams of quadrupedal robots has attracted growing interest for complex tasks in challenging environments. However, physically coupling multiple robots through a shared payload introduces strong dynamical interactions and rigid holonomic constraints, making safe coordinated motion generation under obstacle avoidance particularly challenging.

Distributed predictive control provides a promising framework for cooperative multi-robot transportation by decomposing the coupled robot--payload planning and control problem into local optimization subproblems coordinated through communication and consensus. However, distributed transportation by legged robots remains difficult because the agents must coordinate both local motions and shared transportation variables, including payload trajectories and interaction wrenches, while maintaining consistent distributed predictions. These difficulties are further compounded by the inherent instability of legged robots, unilateral contact constraints, rigid physical coupling, and safety-critical obstacle-avoidance requirements.

To address these challenges, this paper develops a safety-critical distributed nonlinear model predictive control (DNMPC) framework for cooperative payload transportation by teams of quadrupedal robots. The proposed approach decomposes the overall predictive control problem into coordinated local NMPC subproblems while enforcing consensus over shared payload-state and interaction-wrench trajectories through distributed optimization. In addition, acceleration-level holonomic coupling constraints and higher-order control barrier function (HOCBF) safety constraints are incorporated directly within the distributed predictive control formulation.


\vspace{-0.8em}
\subsection{Related Work}
\label{sec:Related_work}

Model predictive control (MPC) has become a powerful framework for planning and control of legged robots due to its ability to systematically handle system dynamics, physical constraints, and task objectives over a finite horizon. Existing MPC approaches span both reduced-order and full-order models. Representative reduced-order models include linear inverted pendulum (LIP)~\cite{kajita19991LIP}, spring-loaded inverted pendulum (SLIP)~\cite{SLIP}, centroidal dynamics~\cite{orin2013centroidal}, and single rigid body (SRB) models~\cite{Kim_Wensing_Convex_MPC_01,Wensing_VBL_HJB,Abhishek_Hae-Won_TRO,Leila_Hamed_RAL,pandala2022robust,Ruturaj_Locomanipulation_RAL}, while full-order nonlinear MPC directly exploits complete robot dynamics for improved performance~\cite{Patrick_TRO_Review,crocoddyl,ProxDDP_TRO}.


\begin{figure*}[t]
    \centering
    \includegraphics[width=\linewidth]{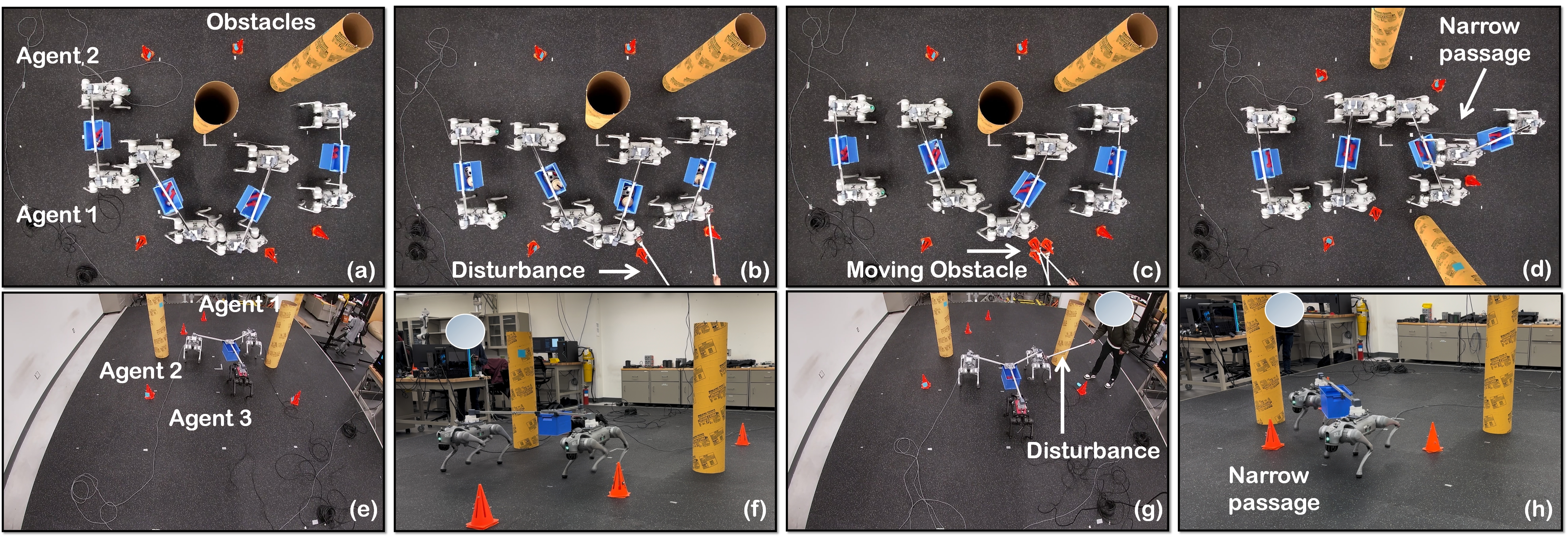}
    \vspace{-1.8em}
    \caption{Experimental snapshots of cooperative payload transportation using Unitree Go2 and A1 robots in a lab environment with conical and cylindrical obstacles. Top row: (a) two-agent nominal transportation, (b) disturbance rejection, (c) moving-obstacle avoidance, and (d) narrow-passage navigation. Bottom row: (e) three-agent nominal transportation, (f) three-agent side view, (g) three-agent disturbance experiment, and (h) side view of the experiment in (d).}
    \vspace{-1.1em}
    \label{fig:snapshots}
\end{figure*}


Safety-critical control has gained significant attention in legged robotics, particularly through the integration of control barrier functions (CBFs) for obstacle avoidance and safe locomotion~\cite{Ames_CBF,CBF_MRS,MRS_CBF_Cavorsi,NMPC_CBF_Sreenath,NMPC_CBF_Ames_Hutter,Basit_RAL,Yicheng_Hamed_IROS}. However, extending such safety-critical predictive control frameworks to cooperative payload transportation remains challenging due to the shared robot--payload dynamics and holonomic interaction constraints.

Recent works have explored predictive control for cooperative quadrupedal coordination and transportation. The works in~\cite{Jeeseop_TRO,Jeeseop_Hamed_ASME,Randy_ICRA_MultiAgent} developed layered distributed and centralized convex MPC frameworks for cooperative locomotion of holonomically constrained quadrupedal robots; however, shared payload dynamics were not explicitly modeled. Distributed MPC for dynamically decoupled quadrupeds was also developed in~\cite{Basit_RAL,Yicheng_Hamed_IROS}, where coordination is achieved primarily through inter-agent safety constraints rather than rigid physical coupling. The work in~\cite{DeVincenti2023CLM} developed a centralized MPC framework for collaborative loco-manipulation using SRB dynamics and multiple-shooting optimization; however, safety-critical constraints were not explicitly incorporated and validation was primarily simulation-based. A safety-critical centralized NMPC framework for cooperative payload transportation was developed in~\cite{Ruturaj_Hamed_2026SafetyCriticalCentralizedNMPC}; however, the resulting optimization problem was solved using a centralized rather than distributed architecture. 

Distributed model predictive control (DMPC) provides a natural framework for coordinating dynamically coupled multi-robot systems by decomposing global optimization problems into local subproblems solved through inter-agent communication~\cite{Ditributed_MPC_Book,2009_Scattolini_DMPC}. Among distributed optimization methods, the alternating direction method of multipliers (ADMM) is particularly attractive due to its ability to decompose coupled problems into parallel local subproblems coordinated through iterative consensus updates~\cite{ADMM_Boyd}.

More recently, ADMM-based distributed MPC approaches have been explored for collaborative transportation and loco-manipulation~\cite{Zhou2026ACLM}. Such approaches typically enforce consensus only over interaction-wrench trajectories while relying on approximate payload-state coupling, without explicitly incorporating rigid holonomic coupling constraints, and have been validated primarily in simulation. In contrast, the proposed framework develops a safety-critical distributed NMPC architecture that enforces consensus over both payload-state and interaction-wrench trajectories while explicitly incorporating acceleration-level holonomic coupling constraints. This enables dynamically consistent payload predictions across agents over the prediction horizon, which is particularly important under rigid coupling and safety-critical obstacle avoidance. The proposed framework is further validated experimentally on teams of quadrupedal robots performing cooperative payload transportation (see Fig.~\ref{fig:snapshots}).

Learning-based approaches for collaborative transportation and loco-manipulation have also been explored~\cite{2509.14342,2411.07104,2603.23278,2509.13239}. In contrast, this work focuses on model-based distributed predictive control with explicit robot--payload dynamics, holonomic coupling, and safety-critical guarantees.


\vspace{-0.8em}
\subsection{Contributions}
\label{sec:Contributions}

The main contributions of this work are threefold. First, a safety-critical distributed nonlinear model predictive control (DNMPC) framework is developed for cooperative payload transportation by teams of quadrupedal robots under rigid holonomic coupling constraints. Second, unlike existing wrench-only ADMM methods, the proposed ADMM-based distributed optimization architecture enforces consensus over both payload-state and interaction-wrench trajectories while explicitly incorporating acceleration-level holonomic coupling constraints and HOCBF-based safety constraints within the distributed predictive control formulation.

Third, the proposed framework is validated through numerical simulations with teams of two, three, and four quadrupedal robots, together with real-time experiments on Unitree Go2 and A1 robots in two- and three-agent cooperative transportation tasks under payload uncertainty, external disturbances, and multiple obstacle configurations (see Fig.~\ref{fig:snapshots}). Comparisons with the corresponding centralized NMPC framework show approximately $9\%$ and $23\%$ reductions in average NLP solve time for the three- and four-agent cases, respectively, while maintaining comparable closed-loop performance. 

Ablation studies further demonstrate robustness to communication delays, quantify the effect of ADMM parameters and iterations, and show that explicit payload-state consensus with holonomic constraints substantially improves payload tracking and distributed coordination over existing wrench-only consensus formulations, reducing the RMS payload yaw tracking error by approximately $4\times$.


\vspace{-0.5em}
\section{Networked Model of Cooperative Transportation}
\label{sec:Networked_Model}

\begin{figure*}[t]
    \centering
    \includegraphics[width=\linewidth]{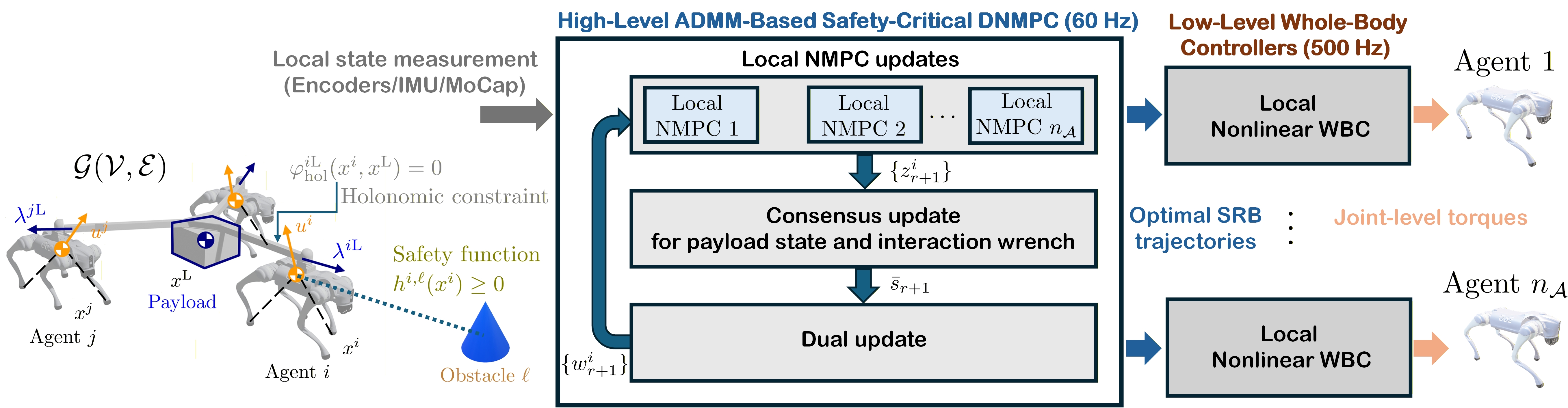}
    \vspace{-1.8em}
    \caption{Overview of the proposed ADMM-based safety-critical distributed NMPC framework for cooperative payload transportation. Parallel local NMPC subproblems \eqref{eq:local_admm_update} are coordinated through ADMM consensus \eqref{eq:consensus_update} and dual updates \eqref{eq:dual_update} to compute optimal state, ground reaction force, and interaction-wrench trajectories, which are tracked by decentralized whole-body controllers.}
    \vspace{-1.2em}
    \label{fig:Overview}
\end{figure*}

This section develops a networked dynamical model for cooperative payload transportation by multiple quadrupedal robots. The proposed framework decomposes the overall predictive control problem into local NMPC subproblems while coordinating the shared payload and interaction trajectories through distributed consensus optimization (see Fig.~\ref{fig:Overview}).


\vspace{-0.75em}
\subsection{Network Structure}
\label{sec:Network_Structure}

We consider a network of $n_{\Aset}$ dynamically coupled robotic agents cooperatively transporting a shared payload. The agents are indexed by $\Aset:=\{1,\dots,n_{\Aset}\}$, and together with the payload node $\Load$, form the vertex set $\Vset:=\Aset\cup\{\Load\}$. The interaction topology is modeled as an undirected star graph $\Graphnet(\Vset,\Eset)$ centered at $\Load$, where each agent interacts only with the payload (see Fig.~\ref{fig:Overview}). Variables associated with node $i\in\Vset$ and edge $\{i,\Load\}\in\Eset$ are denoted by superscripts $i$ and $i\Load$, respectively.


\vspace{-0.75em}
\subsection{Local Dynamics and Coupling Constraints}
\label{sec:Local_Dynamics}

\textit{Local Dynamics:} Each agent $i\in\Aset$ evolves according to the following discrete-time nonlinear dynamics:
\begin{equation}\label{eq:local_dyn}
x^{i}(t+1)
=
f^{i}\big(x^{i}(t),u^{i}(t),\lambda^{i\Load}(t)\big),
\end{equation}
where $t\in\Integer:=\{0,1,\cdots\}$ denotes the discrete-time index, $x^{i}(t)\in\Xset^{i}\subset\Real^{n_{x}}$ is the local state, $u^{i}(t)\in\Uset^{i}\subset\Real^{n_{u}}$ is the local control input, and $\lambda^{i\Load}(t)\in\Real^{n_{\lambda}}$ denotes the interaction wrench exchanged between agent $i$ and the shared payload $\Load$.

Since the payload is physically coupled to all robotic agents through the interaction wrenches, its dynamics depend on the collective interaction forces and torques generated by the robotic team. The payload dynamics are described by
\begin{equation}\label{eq:payload_dyn}
x^{\Load}(t+1)
=
f^{\Load}\big(x^{\Load}(t),\lambda^{\Load}(t)\big),
\end{equation}
where $x^{\Load}(t)\in\Xset^{\Load}\subset\Real^{n_{x}}$ represents the payload state, and $\lambda^{\Load}(t)$ denotes the stack of all interaction wrenches applied to the payload, i.e.,
\begin{equation}\label{eq:payload_wrench_stack}
\lambda^{\Load}(t)
:=
\col(\lambda^{i\Load}(t):i\in\Aset),
\end{equation}
where $\col(\cdot)$ denotes the column stacking operator. Consequently, the payload trajectory depends collectively on the interaction wrenches generated by all robotic agents.

\textit{SRB Model:} Both the robotic agents and the shared payload are modeled using single rigid body (SRB) dynamics. For each agent $i\in\Aset$, the state $x^{i}$ includes the Cartesian position, orientation (Euler angles), and linear and angular velocities, while the control input $u^{i}$ consists of the ground reaction forces (GRFs) subject to friction-cone constraints $\Uset^{i}$. Similarly, $x^{\Load}$ denotes the payload state, and $\lambda^{i\Load}$ represents the interaction wrench exchanged between agent $i$ and the payload.

\textit{Coupling:} The robotic agents are physically connected to the shared payload through rigid coupling mechanisms. Each coupling constrains the relative translation and the relative roll and pitch motions between the corresponding agent and the payload, while allowing free relative yaw motion (see Fig.~\ref{fig:snapshots}). Accordingly, each edge $\{i,\Load\}\in\Eset$ induces a five-dimensional holonomic constraint consisting of three translational constraints and two rotational constraints. The corresponding holonomic constraint is written as
\begin{equation}\label{eq:holonomic_constraints}
    \varphi_{\mathrm{hol}}^{i\Load}\big(x^{i}(t),x^{\Load}(t)\big)=0,
    \quad \forall i\in\Aset.
\end{equation}

Differentiating \eqref{eq:holonomic_constraints} twice along the interconnected dynamics yields the acceleration-level constraint
\begin{equation}\label{eq:acceleration_level_constraints}
    \ddot{\varphi}_{\mathrm{hol}}^{i\Load}\big(
    x^{i}(t),x^{\Load}(t),u^{i}(t),\lambda^{\Load}(t)
    \big)=0,
    \quad \forall i\in\Aset,
\end{equation}
which imposes algebraic constraints on the local control input and the interaction wrenches. In this work, the interaction wrenches are retained as decision variables rather than being eliminated analytically to avoid introducing additional nonlinearities into the distributed optimization problem.


\vspace{-0.75em}
\subsection{Safety Constraints and Control Barrier Functions}
\label{sec:Safety_Constraints}

To ensure safe cooperative transportation in cluttered environments, collision avoidance constraints are imposed for both the robotic agents and the shared payload using higher-order control barrier functions (HOCBFs). Let $\Oset:=\{1,\cdots,n_{\Oset}\}$ denote the set of obstacles, where $o^\ell\in\Real^2$ represents the planar position of obstacle $\ell\in\Oset$.

For each robotic agent and the payload, we define the safety function
\begin{equation}\label{eq:safety_function}
    h^{i,\ell}(x^{i}) := \|g(x^{i})-o^\ell\|-d_{\mathrm{safe}}\geq0,
\end{equation}
for all $i\in\Vset$ and $\ell\in\Oset$, where $g(\cdot)$ maps the SRB state to the planar center of mass (CoM) position and $d_{\mathrm{safe}}>0$ denotes the safety margin (see Fig.~\ref{fig:Overview}).

Since the robotic agents and the payload evolve according to second-order dynamics, the resulting safety constraints have relative degree two. Accordingly, HOCBF constraints are constructed for each agent-obstacle and payload-obstacle pair. The corresponding HOCBF constraints for the robotic agents are given by
\begin{equation}\label{eq:agent_hocbf}
    \psi_{\mathrm{CBF}}^{i,\ell}\big(
    x^{i}(t),
    u^{i}(t),
    \lambda^{i\Load}(t)
    \big)
    \geq 0,
    \quad
    \forall i\in\Aset,\;
    \forall \ell\in\Oset,
\end{equation}
while the payload HOCBF constraints are written as
\begin{equation}\label{eq:payload_hocbf}
    \psi_{\mathrm{CBF}}^{\Load,\ell}\big(
    x^{\Load}(t),
    \lambda^{\Load}(t)
    \big)
    \geq 0,
    \quad
    \forall \ell\in\Oset.
\end{equation}
These safety constraints are enforced directly within the distributed predictive controller. The HOCBF conditions are constructed using the second-order forward difference of the corresponding safety functions; see~\cite{DT-HOCBF} and Section~\ref{sec:Setup and Controller Synthesis} for additional details.


\vspace{-0.5em}
\section{ADMM-Based Safety-Critical DNMPC}
\label{sec:ADMM_DNMPC}

This section develops a distributed nonlinear model predictive control (DNMPC) framework for cooperative payload transportation. Each robotic agent solves a local finite-horizon optimal control problem while maintaining local copies of the shared payload and interaction-wrench trajectories. Consensus among the duplicated shared variables is enforced using the alternating direction method of multipliers (ADMM).


\vspace{-0.75em}
\subsection{Distributed Finite-Horizon Optimal Control Problem}
\label{sec:Distributed_Finite_Horizon_OCP}

We consider a prediction horizon of length $N\in\Integerpos$. For each robotic agent $i\in\Aset$, we define the \textit{local decision variable}
\begin{equation}\label{eq:local_decision_variable}
    z^{i} := \col
    \Big(
    x^{i}(\cdot),
    u^{i}(\cdot),
    x^{\Load,i}(\cdot),
    \lambda^{1\Load,i}(\cdot),
    \cdots,
    \lambda^{n_{\Aset}\Load,i}(\cdot)
    \Big),
\end{equation}
where $x^{i}(\cdot)$ and $u^{i}(\cdot)$ denote the local state and input trajectories over the prediction horizon, respectively. Specifically, $x^{i}(\cdot):=\col\{x^{i}_{t+1|t},\cdots,x^{i}_{t+N|t}\}$ and $u^{i}(\cdot):=\col\{u^{i}_{t|t},\cdots,u^{i}_{t+N-1|t}\}$, where the notation $t+k|t$ denotes the predicted value of a variable at time $t+k$, computed based on information available at time $t$. In addition, $x^{\Load,i}(\cdot):=\col\{x^{\Load,i}_{t+1|t},\cdots,x^{\Load,i}_{t+N|t}\}$ represents the \textit{local copy} of the shared payload trajectory maintained by agent $i$. Similarly, $\lambda^{1\Load,i}(\cdot),\cdots,\lambda^{n_{\Aset}\Load,i}(\cdot)$ denote the local copies of the interaction wrench trajectories maintained by agent $i$. The initial conditions are chosen as the measured system states, i.e., $x^{i}_{t|t}=x^{i}(t)$ and
$x^{\Load,i}_{t|t}=x^{\Load}(t)$.

The local cost function of agent $i$ is defined by
\begin{alignat}{4}
    &J^{i}(z^{i}) &&:= \| x^{i}(\cdot) - x^{i,\reff}(\cdot) \|_{Q^{i}}^{2} + \| u^{i}(\cdot) \|_{R^{i}}^{2} \nonumber\\
    & && \,\,\,+ \| x^{\Load,i}(\cdot) - x^{\Load,\reff}(\cdot) \|_{Q^{\Load}}^{2} + \sum_{j\in\Aset} \| \lambda^{j\Load,i}(\cdot) \|_{R^{\lambda}}^{2},\label{eq:local_cost}
\end{alignat}
which penalizes trajectory-tracking errors, control effort, payload trajectory-tracking objectives, and interaction-wrench magnitudes over the prediction horizon. In our notation, $x^{i,\reff}(\cdot)$ and $x^{\Load,\reff}(\cdot)$ denote the desired agent and payload trajectories, respectively, $Q^{i}\succ0$, $R^{i}\succ0$, $Q^{\Load}\succ0$, and $R^{\lambda}\succ0$ are weighting matrices, and $\|y\|_{Q}^{2}:=y^{\top}Q\,y$.

The \textit{local feasible set} associated with agent $i$ is denoted by $\Zset^{i}$ and consists of: (i) the local SRB dynamics \eqref{eq:local_dyn}, (ii) the payload dynamics \eqref{eq:payload_dyn}, (iii) the holonomic coupling constraints \eqref{eq:acceleration_level_constraints}, (iv) the HOCBF safety constraints \eqref{eq:agent_hocbf} and \eqref{eq:payload_hocbf}, and (v) feasibility constraints over the prediction horizon.

To enable distributed optimization, each robotic agent maintains local copies of the shared payload and interaction wrench trajectories. We define the \textit{local shared trajectory} vector
\begin{equation}\label{eq:shared_trajectory}
    s^{i} :=
    \col\Big(
    x^{\Load,i}(\cdot),
    \lambda^{1\Load,i}(\cdot),
    \cdots,
    \lambda^{n_{\Aset}\Load,i}(\cdot)
    \Big).
\end{equation}
Consensus among the duplicated shared trajectories is enforced through a \textit{global consensus trajectory}
\begin{equation}\label{eq:global_consensus_variable}
    \bar{s}.
\end{equation}

The resulting distributed finite-horizon optimal control problem is formulated as
\begin{align}
\min_{\{z^{i}\},\bar{s}}
\quad & \sum_{i\in\Aset} J^{i}(z^{i}) 
\label{eq:distributed_ocp_cost}
\\
\text{s.t.} \quad &  z^{i}\in\Zset^{i}, \quad \forall i\in\Aset,
\label{eq:distributed_ocp_constraints_1}
\\
& s^{i}=\bar{s}, \quad \,\,\,\forall i\in\Aset.
\label{eq:distributed_ocp_constraints_2}
\end{align}
The local feasibility constraints in \eqref{eq:distributed_ocp_constraints_1} enforce the agent and payload dynamics, holonomic coupling constraints, HOCBF constraints, and feasibility constraints over the prediction horizon.
The equality constraints in \eqref{eq:distributed_ocp_constraints_2} enforce consistency among the local copies of the shared payload and interaction wrench trajectories maintained by the robotic agents.


\vspace{-0.8em}
\subsection{ADMM-Based Distributed Optimization}
\label{sec:ADMM_based_DO}

To solve the distributed optimal control problem \eqref{eq:distributed_ocp_cost}--\eqref{eq:distributed_ocp_constraints_2}, we employ ADMM. Let $w^{i}$ denote the scaled dual variable associated with the consensus constraint $s^{i}=\bar{s}$ for agent $i\in\Aset$. Instead of using a scalar penalty parameter, we use a positive definite block-diagonal penalty matrix $P\succ0$ to account for the different units and relative importance of the payload-state and interaction-wrench trajectories.

The augmented Lagrangian associated with the distributed optimal control problem is given by
\begin{equation}\label{eq:augmented_lagrangian}
\mathcal{L}_{P} := \sum_{i\in\Aset}\left(J^{i}(z^{i})+\frac{1}{2}\left\|s^{i}-\bar{s}+w^{i}\right\|_{P}^{2}\right).
\end{equation}
The quadratic penalty term in \eqref{eq:augmented_lagrangian} penalizes disagreement between the local shared trajectories and the global consensus trajectory, thereby coordinating consistency among the distributed local NMPC problems.

The ADMM algorithm proceeds iteratively by repeatedly updating the local optimization variables, the consensus trajectory, and the dual variables (see Fig.~\ref{fig:Overview}). In the sequel, the subscript $r\in\Integerpos$ denotes the ADMM iteration index associated with the distributed optimization algorithm and should not be confused with the discrete-time index $t$ associated with the system dynamics.

At ADMM iteration $r\in\Integerpos$, the following steps are performed sequentially.

\textit{Step 1 (Local NMPC update):} Each robotic agent solves
\begin{equation}\label{eq:local_admm_update}
z_{r+1}^{i}
=
\argmin_{z^{i}\in\Zset^{i}}
\left(
J^{i}(z^{i})
+
\frac{1}{2}
\left\|
s^{i}-\bar{s}_{r}+w_{r}^{i}
\right\|_{P}^{2}
\right),
\quad
\forall i\in\Aset.
\end{equation}
This step computes a locally feasible trajectory satisfying the agent dynamics, payload dynamics, holonomic constraints, HOCBF constraints, and feasibility constraints encoded in $\Zset^{i}$, while penalizing disagreement with the current consensus trajectory $\bar{s}_{r}$.

\textit{Step 2 (Consensus update):} After the local solves, the shared trajectory is updated by averaging the shifted local copies:
\begin{equation}\label{eq:consensus_update}
\bar{s}_{r+1}
=
\frac{1}{n_{\Aset}}
\sum_{i\in\Aset}
\left(
s_{r+1}^{i}
+
w_{r}^{i}
\right).
\end{equation}
The consensus update minimizes \eqref{eq:augmented_lagrangian} with respect to the shared trajectory $\bar{s}$ while keeping the local and dual variables fixed, yielding the common payload and interaction-wrench trajectory that best reconciles all local predictions.

\textit{Step 3 (Dual update):} Finally, the scaled dual variables are updated as
\begin{equation}\label{eq:dual_update}
w_{r+1}^{i}
=
w_{r}^{i}
+
s_{r+1}^{i}
-
\bar{s}_{r+1},
\quad
\forall i\in\Aset.
\end{equation}
This update accumulates the consensus mismatch and drives the duplicated shared trajectories toward agreement.

\textit{Iteration:} The three ADMM steps above are repeated iteratively over a finite number of ADMM iterations to coordinate agreement among the duplicated shared trajectories across the distributed local NMPC problems. The local optimization problems in \eqref{eq:local_admm_update} are independent across agents and can therefore be solved simultaneously in parallel. Consequently, for a team of $n_{\Aset}$ robotic agents, the proposed formulation results in $n_{\Aset}$ local NMPC subproblems at each ADMM iteration.

To monitor consensus convergence across ADMM iterations, define the primal residual
\begin{equation}\label{eq:primal_residual}
r_{\mathrm{prim},r}^{i}
:=
s_{r}^{i}
-
\bar{s}_{r},
\quad
\forall i\in\Aset,
\end{equation}
and the dual residual $r_{\mathrm{dual},r}:= P\left(\bar{s}_{r} - \bar{s}_{r-1}\right)$, which measure the disagreement between the local and consensus trajectories and the change in the consensus trajectory, respectively.

\textit{Implementation:} The proposed controller is implemented in a receding-horizon fashion. At each sampling instant, the local NMPC subproblems are warm-started using the shifted previous solution, followed by a fixed number of ADMM iterations before applying the first control input. Numerical implementation details are given in Section~\ref{sec:Experiments}.

\begin{remark}
Unlike the formulation in~\cite{Zhou2026ACLM}, which introduces a separate payload NMPC subproblem, the proposed framework maintains local copies of the payload-state and interaction-wrench trajectories within each agent's NMPC, resulting in only $n_{\mathcal{A}}$ parallel local NMPC subproblems rather than $n_{\mathcal{A}}+1$. Introducing a separate payload subproblem while enforcing consensus over both payload-state and interaction-wrench trajectories would require additional nonlinear consensus coordination associated with the holonomic coupling constraints, thereby increasing ADMM complexity.
\end{remark}


\vspace{-0.5em}
\section{Experiments}
\label{sec:Experiments}

This section evaluates the proposed ADMM-based safety-critical DNMPC framework through numerical simulations on two-, three-, and four-agent quadrupedal teams and real-time experiments on two- and three-agent teams performing cooperative payload transportation in cluttered environments.


\vspace{-0.75em}
\subsection{Setup and Controller Synthesis}
\label{sec:Setup and Controller Synthesis}

This work employs Unitree Go2 quadrupedal robots for numerical simulations and hardware experiments. Since only two Go2 robots are available experimentally, the three-agent experiments use a heterogeneous team comprising two Go2 robots and one Unitree A1 robot. Each Go2 robot has an 18-DoF floating-base configuration, consisting of 12 actuated joints and a 6-DoF torso, with a mass of approximately 15~kg and a nominal standing height of 0.28~m.

\textit{Layered Control:} The control framework is implemented on an offboard desktop computer (Intel i9-14900KF, 64~GB DDR5). The high-level ADMM-based safety-critical DNMPC runs at 60~Hz with one parallel thread per local NMPC subproblem. Decentralized nonlinear whole-body controllers (WBCs), adopted from~\cite{Randy_Paper_LCSS}, run at 500~Hz with one dedicated thread per robot to generate torque commands (see Fig.~\ref{fig:Overview}). Although implemented on a multithreaded single computer for synchronization and reproducibility, the proposed optimization architecture is inherently distributed and readily extendable to multi-computer deployments.

\textit{Estimation:} Robot states are measured using joint encoders, onboard IMUs, and kinematic estimators. To eliminate drift in onboard state estimation, all experiments are conducted in a motion-capture (MoCap) environment, where the robot positions and obstacle locations are obtained using a six-camera OptiTrack system. The payload SRB state is reconstructed using a kinematic estimator based on the robot states and rigid coupling geometry, with payload yaw determined from the relative agent positions due to the yaw-free coupling.

\textit{NMPC Hyperparameters:} Both the robotic agents and the payload are modeled using SRB dynamics with $n_x=12$ states. Each robot has $n_u=12$ control inputs corresponding to the GRFs, and each interaction wrench has dimension $n_{\lambda}=6$. Each local NMPC employs a prediction horizon of $N=8$ and a sampling period of $T_s=16.7$~ms (60~Hz). A nominal trot gait is considered for each robot, with contact schedules and foothold locations generated using Raibert's heuristics~\cite{raibert1986legged} embedded within the low-level WBC. The friction coefficient is set to $\mu=0.4$, and the normal contact forces are constrained by $0 \leq f_z \leq 250$~N. The SRB stage weighting matrix for each robot and the payload is selected as $Q^{\SRB}=\bdiag\{Q^{\SRB}_{p},Q^{\SRB}_{\dot p},Q^{\SRB}_{\theta},Q^{\SRB}_{\omega}\}$, with $Q^{\SRB}_{p}=\diag\{1\ex7,1\ex7,16\ex7\}$, $Q^{\SRB}_{\dot p}=1\ex6\,\identity_{3}$, $Q^{\SRB}_{\theta}=1\ex7\,\identity_{3}$, and $Q^{\SRB}_{\omega}=1\ex5\,\identity_{3}$. Here, $p$, $\dot p$, $\theta$, and $\omega$ denote the CoM position, linear velocity, XYZ Euler angles, and angular velocity, respectively, expressed in the world frame. The terminal weighting matrix is chosen as $P^{\SRB}=10\,Q^{\SRB}$, while the contact-force penalty matrix is set to $R^{\SRB}=20\,\identity_{12\times12}$. The interaction-wrench weighting matrix is selected as $R^{\lambda}=\bdiag\{R_f,R_{\tau}\}=\bdiag\{50\,\identity_{3\times3}, 500\,\identity_{3\times3}\}$.

For each safety function $h_{t}^{i,\ell}$ in \eqref{eq:safety_function}, relative-degree-two HOCBF constraints \eqref{eq:agent_hocbf}--\eqref{eq:payload_hocbf} are constructed using the second-order forward-difference formulation in \cite{DT-HOCBF} as
\begin{equation}
    h_{t+2}^{i,\ell}-(2-\alpha_{1}-\alpha_{2})\,h_{t+1}^{i,\ell}
    +(1-\alpha_{1})(1-\alpha_{2})\,h_{t}^{i,\ell} \geq 0,
\end{equation}
and enforced over the NMPC control horizon with $(\alpha_{1},\alpha_{2})=(0.4,0.04)$ and a safety margin of $d_{\mathrm{safe}}=0.6$~m.

\textit{ADMM Parameters and Real-Time Computation:} The ADMM decomposition splits the cooperative NMPC into $n_{\Aset}$ robot--payload local NLPs. Each local NLP includes one robot SRB trajectory, a local payload copy, contact-force trajectories, and local interaction-wrench copies. Consensus is enforced over the payload trajectory and interaction wrenches using the penalty matrix $P=\bdiag\{\rho_x\identity,\rho_\lambda\identity\}$ with $\rho_x=1\ex4$ and $\rho_\lambda=1\ex3$. The local NLPs are formulated in CasADi~\cite{CasADI} and solved in parallel using IPOPT~\cite{IPOPT}. At each NMPC update, two ADMM iterations are performed (see Section~\ref{sec:Ablation Studies and Comparisons} for ablation studies on the penalty matrix and number of iterations), and each IPOPT solve is warm-started using the shifted solution from the previous step.

\textit{Comparison with Centralized NMPC:}
Each local NMPC subproblem in \eqref{eq:local_admm_update} contains $N(2n_x+n_u+n_{\Aset}n_{\lambda})+2n_x$ decision variables (including initial conditions in CasADi), whereas the corresponding centralized formulation in \cite{Ruturaj_Hamed_2026SafetyCriticalCentralizedNMPC} contains $(n_{\Aset}+1)(N+1)n_x+n_{\Aset}N(n_u+n_{\lambda})$. Table~\ref{tab:solve_time} compares the mean and standard deviation of NLP solve times (ms) for the proposed distributed framework and the centralized NMPC formulation in~\cite{Ruturaj_Hamed_2026SafetyCriticalCentralizedNMPC} for two-, three-, and four-agent teams in RaiSim \cite{RAISIM}. Both formulations satisfy the 16.7~ms MPC sampling requirement. For the two-agent case, the distributed formulation exhibits slightly higher solve time due to ADMM coordination overhead and implementation-level thread allocation in IPOPT. As the team size increases, however, the proposed distributed framework reduces the average NLP solve time by approximately $9\%$ and $23\%$ for the three- and four-agent cases, respectively, while maintaining comparable closed-loop performance, thereby providing additional computational margin for real-time deployment.


\begin{table}[t]
\caption{Comparison of distributed and centralized NMPC formulations.}
\vspace{-1em}
\label{tab:solve_time}
\centering
\setlength{\tabcolsep}{3pt}
\renewcommand{\arraystretch}{1.05}
\begin{tabular}{c|cc|c|cc|cc}
\toprule
&
\multicolumn{2}{c|}{Decision Variables}
&
\multicolumn{1}{c|}{Iterations}
&
\multicolumn{2}{c|}{Distributed (ms)}
&
\multicolumn{2}{c}{Centralized (ms)}
\\
$n_{\Aset}$
& Dist. & Cent.
& IPOPT
& Avg. & Std.
& Avg. & Std.
\\
\midrule
2 & 408 & 612  & 6 & 9.28 & 0.27 & 7.62 & 0.66 \\
3 & 456 & 864  & 6 & 11.34 & 0.97 & 12.46 & 1.11 \\
4 & 504 & 1116 & 5 & 11.55 & 0.82 & 15.07 & 0.95 \\
\bottomrule
\end{tabular}
\vspace{-2em}
\end{table}


\vspace{-1em}
\subsection{Hardware Experiments}
\label{sec:Hardware Experiments and Simulations}

Experimental validation is conducted using two robotic teams: (i) a two-agent team of Unitree Go2 robots connected via a custom rigid coupling mechanism with a central basket, and (ii) a heterogeneous three-agent team comprising two Unitree Go2 robots and one Unitree A1 robot connected via a star-shaped coupling mechanism (see Fig.~\ref{fig:snapshots}). Both mechanisms constrain translational and roll/pitch relative motions while allowing free relative yaw motion. Experiments are performed with a nominal 5~kg payload (33.33\% of a single robot's mass) and under payload uncertainty with masses up to 12~kg (80\%), as well as under unknown external push disturbances. All experiments are conducted in a cluttered laboratory environment with up to seven obstacles.

\begin{figure}
    \centering
    \includegraphics[width=\linewidth]{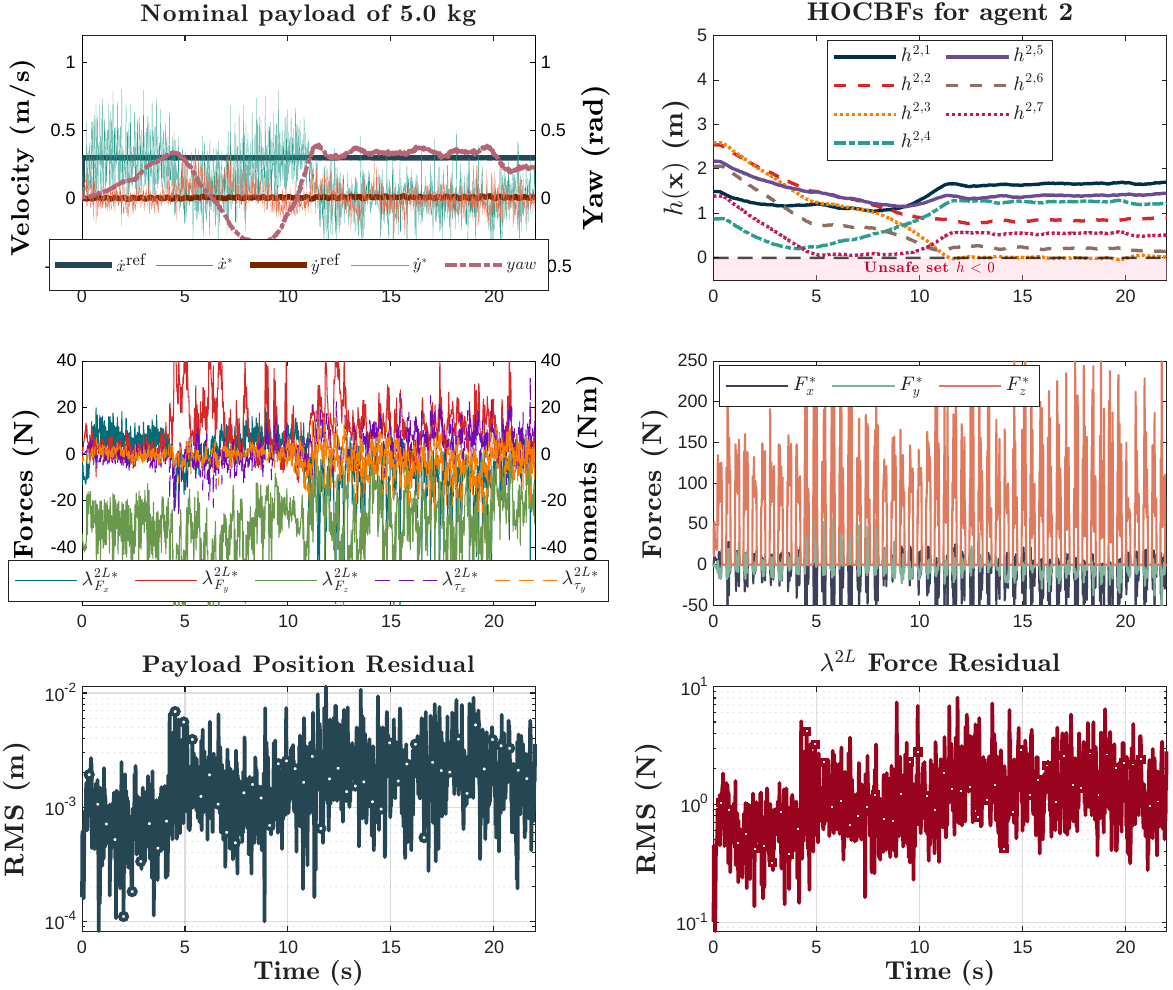}
    \vspace{-2em}
    \caption{Nominal two-agent cooperative transportation with a 5~kg payload in a cluttered environment with $n_{\Oset}=7$ obstacles. The plots show the measured CoM velocity of agent~2, payload yaw, HOCBF safety functions, optimal interaction wrenches and GRFs, and RMS ADMM residuals for payload-position and interaction-wrench consensus. The controller maintains safe tracking and stable consensus, with minor apparent safety-boundary violations due to MoCap noise.}
    \vspace{-1.0em}
    \label{fig:two_agent_nominal_exp}
\end{figure}

\begin{figure}
    \centering
    \includegraphics[width=\linewidth]{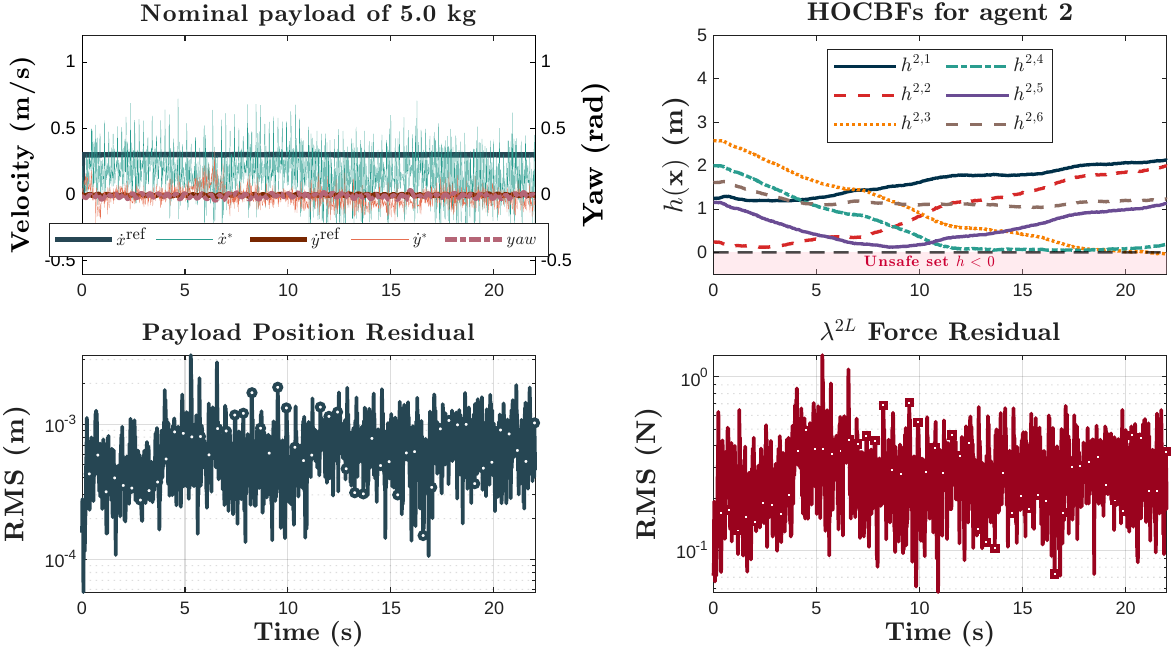}
    \vspace{-2em}
    \caption{Nominal three-agent cooperative transportation with a 5~kg payload in a cluttered environment with $n_{\Oset}=6$ obstacles. The plots show the measured CoM velocity of agent~2, payload yaw, HOCBF safety functions, and RMS ADMM residuals. The controller maintains safe transportation and stable consensus among the local NMPC subproblems, with minor apparent safety-boundary violations attributed to MoCap measurement noise.}
    \vspace{-1.3em}
    \label{fig:three_agent_nominal_exp}
\end{figure}

\textit{Nominal Experiments:}
Reference trajectories are assigned to both the robotic agents and the shared payload at a nominal transportation speed of 0.3~m/s. The proposed controller maintains closed-loop stability of the interconnected robot--payload system while ensuring safe trajectory tracking and collision avoidance (see Fig.~\ref{fig:snapshots}). Figure~\ref{fig:two_agent_nominal_exp} shows the nominal two-agent experiment with a 5~kg payload. The controller autonomously adjusts the payload orientation to maneuver through cluttered obstacles while maintaining stable ADMM consensus, as evidenced by the convergent primal residuals in \eqref{eq:primal_residual} for payload-state and interaction-wrench consensus. A slight apparent violation of two HOCBF safety functions is observed over a brief interval, which is attributed to MoCap measurement noise rather than an actual safety breach. Figure~\ref{fig:three_agent_nominal_exp} shows the three-agent experiment, exhibiting similar safe cooperative transportation and stable consensus behavior.

\begin{figure}
    \centering
    \includegraphics[width=\linewidth]{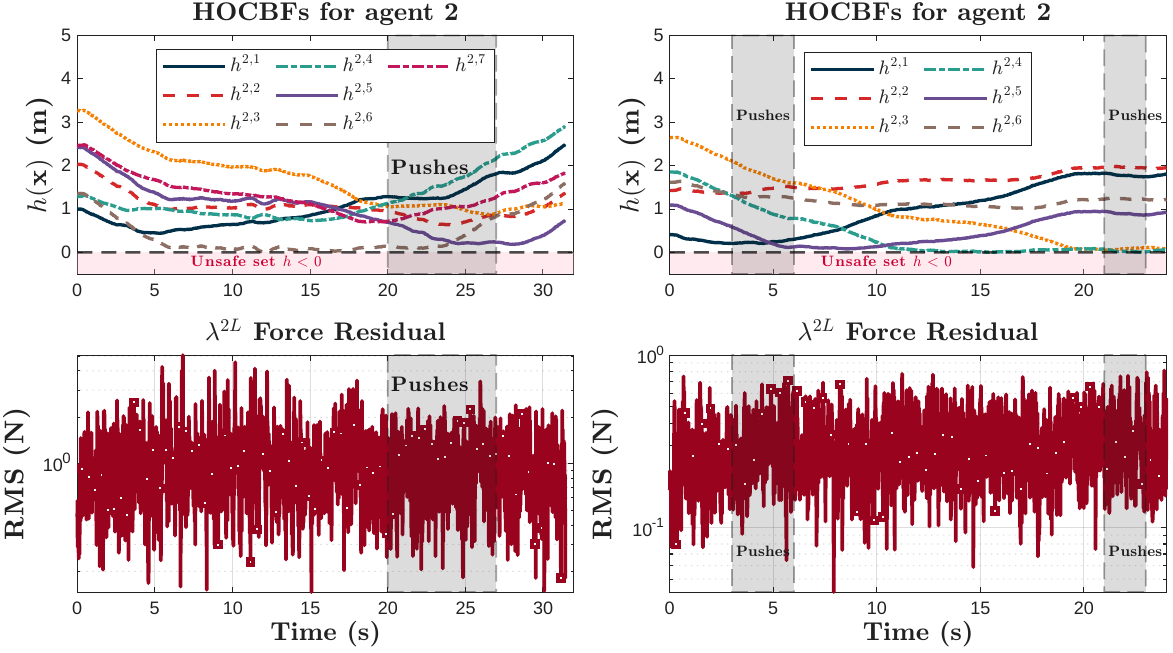}
    \vspace{-2.1em}
    \caption{Robustness experiments under payload uncertainty and external disturbances. Left: two-agent transportation with an unknown 12~kg payload. Right: three-agent transportation with a nominal 5~kg payload. The plots show HOCBF safety functions and RMS ADMM force residuals. External pushes may briefly violate the safety boundary, but the controller rapidly recovers while maintaining stable transportation.}
    \vspace{-1.4em}
    \label{fig:robustness}
\end{figure}

\textit{Robustness to Payload Uncertainty and Disturbances:}
Additional experiments are performed under payload uncertainty and external disturbances using the same setup. Besides the unmodeled increase in payload mass, a bystander applies external pushes to one or more robotic agents (see Fig.~\ref{fig:snapshots}). Figure~\ref{fig:robustness} shows the HOCBF safety functions and ADMM residuals for the two- and three-agent robustness experiments. Despite these disturbances, the proposed controller maintains stable cooperative transportation and robust tracking. Strong pushes may induce brief apparent safety-boundary violations, but the controller rapidly recovers and drives the system back to the safe set while maintaining convergent consensus.

\begin{figure}
    \centering
    \includegraphics[width=1.0\linewidth]{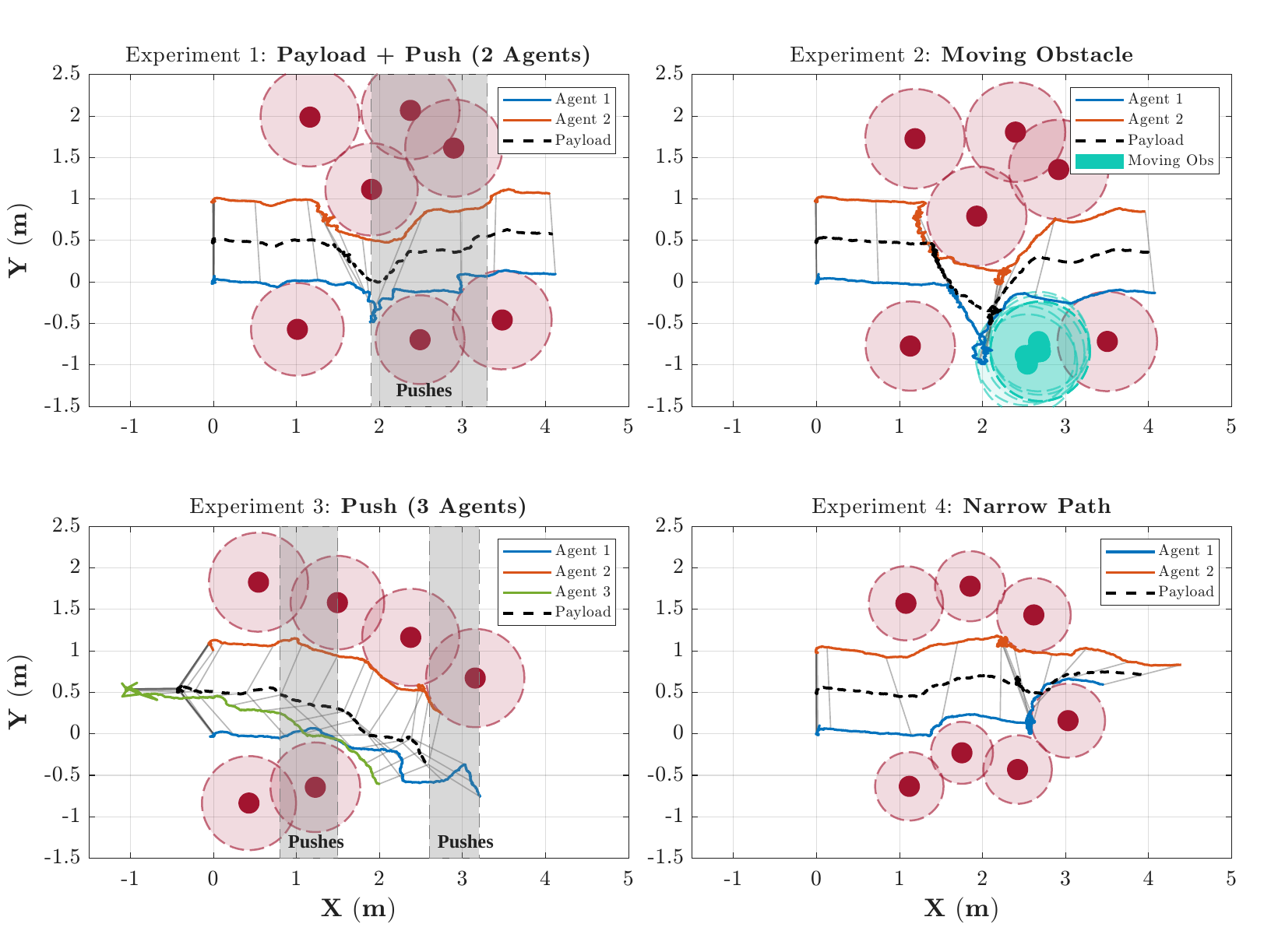}
    \vspace{-2.6em}
    \caption{CoM trajectories of the robotic agents and shared payload in the $xy$-plane. Gray lines indicate rigid coupling; solid circles and dashed regions denote obstacles and safe sets. Left: robustness experiments in Fig.~\ref{fig:robustness}. Right top: two-agent experiment with a moving obstacle. Right bottom: two-agent experiment in a narrow passage with $d_{\mathrm{safe}}=0.45$~m.}
    \vspace{-1.7em}
    \label{fig:xy_traj}
\end{figure}

\textit{Additional Obstacle Configurations:}
Figure~\ref{fig:xy_traj} shows the CoM trajectories of the robotic agents and payload in the $xy$-plane for additional obstacle configurations, including a moving obstacle and a narrow passage with $d_{\mathrm{safe}}=0.45$~m (see Fig.~\ref{fig:snapshots}), further demonstrating safe navigation and cooperative transportation in tightly constrained environments. Minor trajectory jitter is observed in the moving-obstacle experiment, which is primarily attributed to MoCap measurement noise.


\vspace{-0.8em}
\subsection{Ablation Studies and Comparisons}
\label{sec:Ablation Studies and Comparisons}

\begin{figure}
    \centering
    \includegraphics[width=\linewidth]{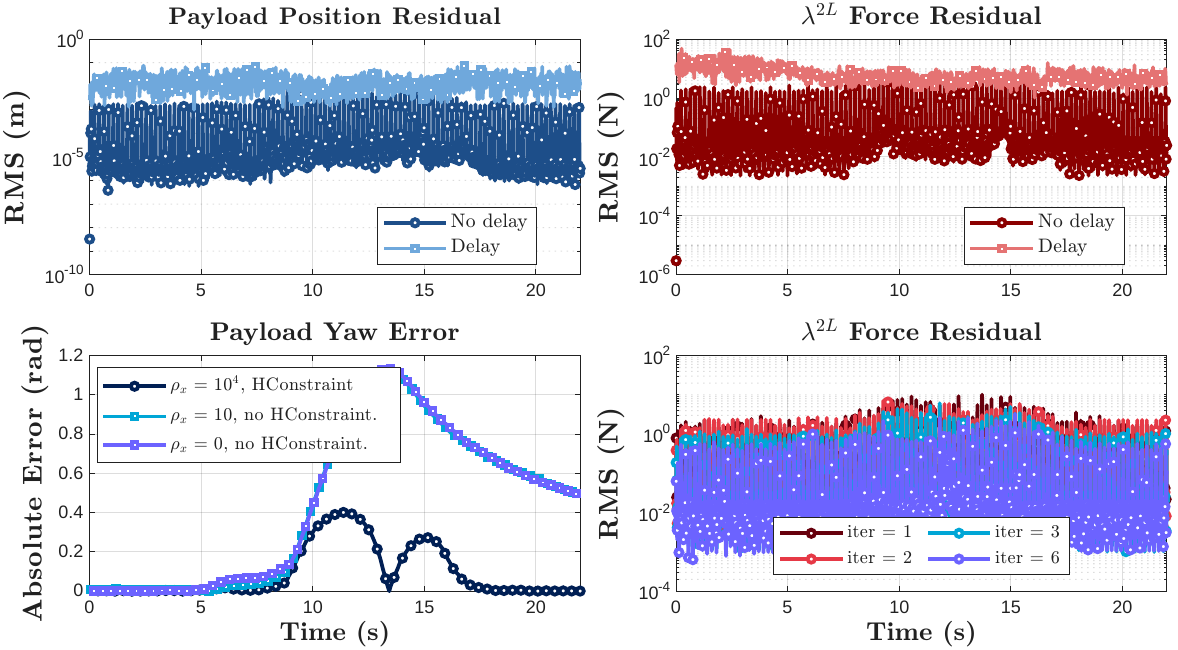}
    \vspace{-2em}
    \caption{Ablation studies on communication delay, ADMM consensus gains, and ADMM iterations. Top row: RMS payload-position (left) and interaction-force (right) residuals under nominal and delayed communication. Bottom left: Absolute payload yaw tracking error for the nominal setting ($\rho_x=10^4$ with holonomic constraints) and reduced state-consensus gains ($\rho_x=10^1$ and $\rho_x=0$) without holonomic constraints. Bottom right: RMS interaction-force residuals for ADMM iterations ${1,2,3,6}$.}
    \vspace{-1.2em}
    \label{fig:ablation_comparision}
\end{figure}

\textit{1) Sensitivity to ADMM Communication Delay:}
In the current implementation, ADMM runs in a multithreaded fashion on a single computer. To emulate delays arising in future multi-computer networked implementations, stochastic bidirectional communication delays and packet loss are introduced during ADMM message exchange while keeping the MPC rate, horizon, and ADMM iterations fixed. Messages exchanged between the coordinator and robotic agents experience random delays of up to 20 communication rounds (approximately ten MPC updates) with independent packet dropouts of probability $p=0.5$, handled using a hold-last policy. Figure~\ref{fig:ablation_comparision} (top row) shows that communication delays increase payload-state and interaction-force residual oscillations, but the controller maintains stable transportation and convergent consensus.


\textit{2) Sensitivity to Consensus Gains and Holonomic Constraints:}
To evaluate the importance of explicit payload-state consensus, we compare the nominal setting ($\rho_x=10^4$ with holonomic constraints) against reduced state-consensus gains ($\rho_x=10^1$ and $\rho_x=0$) without holonomic constraints to approximate wrench-only ADMM formulations such as~\cite{Zhou2026ACLM}. Figure~\ref{fig:ablation_comparision} (bottom left) shows that weakening payload-state consensus significantly degrades payload yaw tracking during cooperative transportation. As summarized in Table~\ref{tab:ablation_summary}, the proposed framework achieves an RMS payload yaw tracking error of $1.16\times10^{-1}$, reducing the tracking error by approximately $4\times$ compared with the reduced-gain cases ($4.81\times10^{-1}$ and $4.82\times10^{-1}$). These results demonstrate that explicit payload-state consensus and holonomic coupling are essential for accurate payload tracking and distributed coordination.


\textit{3) Sensitivity to ADMM Iterations:} To evaluate the trade-off between consensus quality and computational cost, we compare ADMM iterations of ${1,2,3,6}$ per NMPC update while keeping all other parameters fixed. Figure~\ref{fig:ablation_comparision} (bottom right) shows that increasing the number of ADMM iterations reduces interaction-force residuals and improves consensus quality. While a single iteration yields noticeably larger residual oscillations, two ADMM iterations provide sufficient consensus quality for stable cooperative transportation while satisfying the real-time computational requirements of the proposed controller (see Table~\ref{tab:ablation_summary}).



\begin{table}[t]
\centering
\caption{Time-averaged RMS metrics for the ablation studies in Fig.~\ref{fig:ablation_comparision}.}
\label{tab:ablation_summary}
\vspace{-1em}
\scriptsize
\setlength{\tabcolsep}{3pt}
\begin{tabular}{lcc}
\toprule
Study & Metric & Avg. RMS \\
\midrule

Comm. delay
& Pos. residual (nom. / delay)
& $6.97\times10^{-5}/1.87\times10^{-2}$ \\

& Force residual (nom. / delay)
& $7.40\times10^{-2}/5.93$ \\

Cons. \& HCs
&
\begin{tabular}[c]{@{}c@{}}
Payload yaw error ($\times10^{-1}$)\\
\footnotesize (\textbf{ours} / red. state consensus)
\end{tabular}
&
$\mathbf{1.16}\;(\downarrow\mathbf{4\times}),\,4.81,\,4.82$ \\

ADMM iter.
& Force residual (iter 1,2,3,6) ($\times10^{-2}$)
& $9.93,\,7.70,\,4.92,\,2.62$ \\

\bottomrule
\end{tabular}
\vspace{-2.0em}
\end{table}


\vspace{-0.5em}
\section{Conclusions}
\label{sec:Conclusions}

This work presented an ADMM-based safety-critical distributed NMPC framework for cooperative payload transportation by multi-quadrupedal robotic systems under rigid coupling constraints. Unlike existing wrench-only ADMM methods, the proposed formulation enforces consensus over both payload-state and interaction-wrench trajectories while explicitly incorporating holonomic coupling and HOCBF-based safety constraints. Numerical simulations with two, three, and four robotic agents, together with real-time experiments on two- and three-agent teams, demonstrated safe cooperative transportation in cluttered environments under payload uncertainty and external disturbances.

For the three- and four-agent cases, comparisons with the corresponding centralized NMPC formulation showed approximately $9\%$ and $23\%$ reductions in average NLP solve time, respectively, while maintaining comparable closed-loop transportation performance. Ablation studies showed that communication delays primarily degrade force consensus, explicit payload-state consensus with holonomic constraints improves payload tracking and distributed coordination by reducing the RMS payload yaw tracking error by approximately $4\times$ compared with wrench-only formulations, and two ADMM iterations provide an effective balance between consensus quality and real-time computation. Although formal convergence guarantees for finite-iteration nonconvex ADMM remain open, the residual analysis and hardware experiments demonstrated sufficient consensus for real-time safe control. 

While large-scale robot teams were not the primary focus of this work, future research will investigate fully decentralized multi-computer implementations with asynchronous communication, richer environment representations, and integration with long-horizon kinematic planners for more complex cooperative maneuvers.




\vspace{-0.5em}
\bibliographystyle{IEEEtran}
\bibliography{references}

\end{document}